# Task Tree Retrieval for Robotic Cooking

Sandeep Bondalapati

*Abstract*— **Robotics is used to foster creativity. Humans can perform jobs in their unique manner, depending on the circumstances. This situation applies to food cooking. Robotic technology in the kitchen can speed up the process and reduce its workload. However, the potential of robotics in the kitchen is still unrealized. In this essay, the idea of FOON, a structural knowledge representation built on insights from human manipulations, is introduced. To reduce the failure rate and ensure that the task is effectively completed, three different algorithms have been implemented where weighted values have been assigned to the manipulations depending on the success rates of motion. This knowledge representation was created using videos of open-sourced recipes.**

## I. INTRODUCTION

Developing intelligent systems that can mimic human behavior and perform tasks needs a huge knowledge-learning dataset to train the machine [1]. The ability of robots depends on the programming model implemented for knowledge retrieval with a less redundant and efficient data set. Based on previous work, the functional object-oriented network [2], [3] is designed with the concept of object-action representation in mind, which describes the interaction between objects and manipulation actions as network nodes and edges. It gives the relationship between objects and their associated manipulations. It has basic units called functional units, which represent individual sub-tasks present in the whole network. Each functional unit consists of objects and their states before and after motion. This basic unit can be formed by the collection of a sequence of actions, which is a task tree, and multiple task tree combinations will give us the outcome.

The FOON is created from the video annotations of various YouTube videos [1], and there can be a chance of making the same recipe in multiple ways with the same and different ingredients available in the same and different states. Objects can have different sizes, shapes, and compositions [3]. In this scenario, the robot should add and avoid certain ingredients in the reference task tree based on the required recipe. I have implemented three different algorithms considering heuristic conditions and level-wise task tree searching for extracting modified reference task trees to match the set of ingredients for a given recipe. This is based on the notion that if we can manipulate one set of things, we can also manipulate similar ones. From a different perspective, if we lack a specific object, we can use objects that are similar to it to complete the task. With the use of these ideas, we can create compressed and expanded networks based on item categories that have information beyond what a conventional network would have.

This paper is organized as follows: in Section 2, the FOON structure with video annotation is described, which is followed by Section 3, where the task retrieval algorithms are explained briefly. In Section 4, the performance of all algorithms is compared in terms of space and time complexity.

## II. FUNCTIONAL OBJECT-ORIENTED NETWORK

The functional object-oriented network is a knowledge representation that is used to serve the robot to do a task. It has been created from several YouTube video sources. FOON is a bipartite network consisting of two types of nodes: object nodes and motion nodes. The object nodes can be the inputs and outputs of a functional unit. Edges connect the objects to motion nodes, enforcing action sequencing. No two object nodes or two motion nodes cannot connect by the edge. The state of the object changes after the motion.

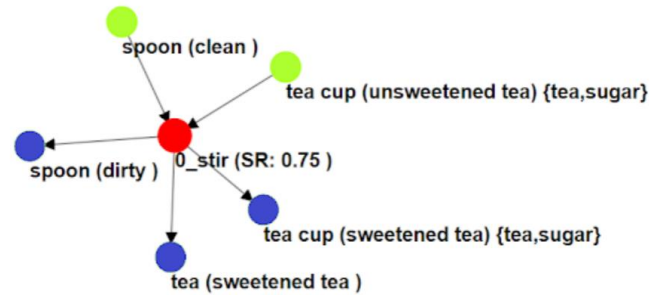

Fig. 1: A basic functional unit with two input nodes and three output nodes connected by an intermediary single-motion node

*Functional Unit:* The whole unit where input and output objects coupled with motion node referred is referred to as a functional unit. The green object nodes associated with edges pointing to the functional motion node are known as input object nodes, and the blue object nodes connected with edges heading away from the functional red color motion node are known as output object nodes, as shown in Figure 1.

*Subgraph:* A subgraph represents an activity; it contains multiple functional units in a sequence where the output nodes of the previous unit act as input nodes to the current functional unit. Multiple subgraphs are combined to form a universal FOON network. This process can be done by a union operation on all the functional units of the subgraph by removing duplicate units.

*Task tree:* A FOON is used to represent knowledge that can be used by the robot for problem-solving. For a given goal, using task retrieval, a robot can fetch a subgraph that contains the sequence of functional units to achieve the respective task.

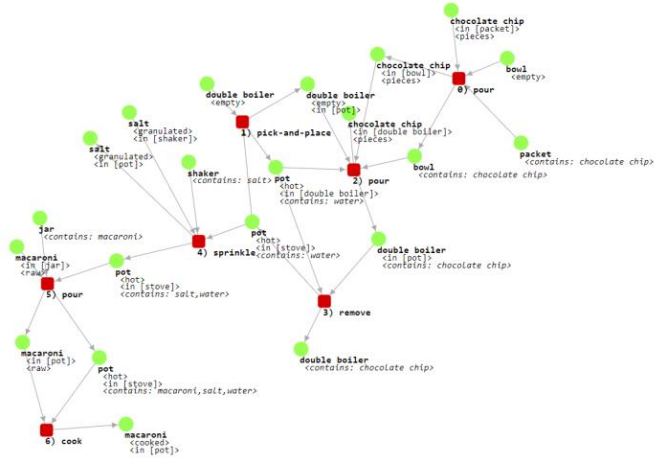

Fig. 2: A Task tree showing the process of making macaroni.

A FOON is created by using the annotations of video demonstrations from YouTube, Activity-Net [11], and Epic-KITCHENS by converting them into a graph structure.

| // | | |
|---|---|---|
| O | tea cup | 0 |
| S | unsweetened tea | { tea, sugar } |
| O | spoon | 1 |
| S | clean | |
| M | stir | Assumed |
| O | tea | 0 |
| S | sweetened tea | |
| O | tea cup | 0 |
| S | sweetened tea | { tea, sugar } |
| O | spoon | 1 |
| S | dirty | |
| // | | |

Fig. 3: A Sample annotation of making sweetened tea

Figure 3 shows the annotation file for the task tree shown in Figure 1. O denotes the object; S denotes the state of the object and M denotes the Motion. Objects can be in moving or in an ideal state denoted by "1" or "0". An object can have multiple ingredients which are denoted in the state, those ingredients can be objects of the functional units.

Figure 2 has 7 functional units each consisting of one motion node and the outcomes of each sequential functional unit act as inputs for the next coming functional unit as shown. Each functional unit is separated by "//" from one other. In order to improve the efficiency of the robot's work, weights are given to each functional unit corresponding to its success rate of execution by the robot [6]. These success rates are based on not only the manipulation but also depends on the type of objects in the functional unit. These values can be based on 1) the physical strength of the robot 2) past experiences in task execution 3) tools or objects to manipulate [4]. These values may vary between 0 to 1 for each type of robot, so they must be carefully analyzed.

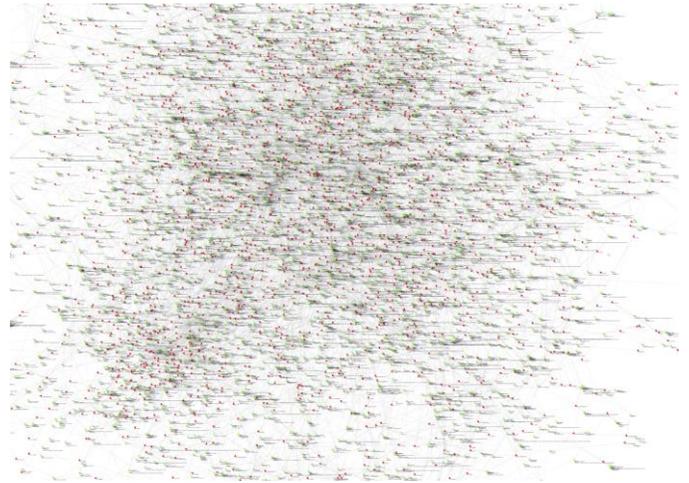

Fig. 4: A Universal FOON network merged with multiple subgraphs.

### III. METHODOLOGY

From a universal FOON, a recipe's task tree having all functional units related to the final state of a dish which is referred to as a goal node has been extracted by using the below algorithms.

*Iterative deepening search:*

This algorithm combines the benefits of Breadth-first search and Depth-first search. Unlike DFS it will do a deepening search level by level in the depth bound by increasing the depth until the solution is found.

1) If the goal node is present in the kitchen no search happens.

2) Otherwise, the goal node presence will be checked in the universal FOON. If it has the goal node, then one of the functional units is selected at Depth 1.

3) The input nodes of each functional unit will be explored and at every level, they will be checked in the kitchen. A task tree is considered a solution if the leaf nodes are available in the kitchen. This condition is assumed as Depth==0. The same nodes are traversed multiple times for every new increment in-depth bound. But it finds the goal task tree at the upper level. If a solution lies at a deeper level this algorithm will take more time compared to the DFS.

4) A separate list containing Boolean values is used to track the condition of a goal node. If the leaf nodes at Depth 0 are present

in the kitchen, then the True value will be added to the list otherwise False will be added. After traversing the $0^{th}$ depth all the list elements will be checked and assigned the value to the flag variable. If the task tree for the goal node is found, then the flag value is True otherwise it is False.

---
Algorithm 1: Task Tree retrieval for the goal nod
---

**Input**: reference goal node $Gn$ and given ingredients $I$
$T \leftarrow$ A list of functional units in the task tree
$K \leftarrow$ A list of items available in the kitchen
$fl \leftarrow$ flag list, to keep track of leaf nodes in kitchen
**If** Depth is 0 **then**
  **if** $Gn$ in $K$ **then**
    $fl$.append(True)
  **else**
    $fl$.append(False)
**end if**
**if** Depth >0 **then**
  **if** $Gn$ does not exist in $K$ **then**
    $C \leftarrow$ Find all functional units that create $L$
    $C' \leftarrow$ Select one from $C$
    $T$.append($C'$)
    **for** each input n in $C'$ **do**
      **if** n is not visited **then**
        Mark n as visited
        Search_IDDFS(n, Depth-1, $T$)
      **end if**
    **end for**
  **end if**
flag=True $\leftarrow$ If all elements in fl is True else False
**if** flag == True **then**
  return $T$.reverse()
**Output**: $T$

*Heuristic 1:*
    This search algorithm is similar to the Breadth-first search, where the search happens level-wise. While selecting functional units at every level instead of selecting anyone the functional unit with the highest success rate of manipulation will be selected. This can be achieved by comparing all the unit's success rates. Here weighted values are considered for the comparison.

---
Algorithm 2: Task tree retrieval using heuristic 1
---

**Input**: reference goal node $Gn$ and given ingredients $In$
$T \leftarrow$ A list of functional units in the task tree
$Kit \leftarrow$ A list of items available in the kitchen
$Qu \leftarrow$ A queue of items to search
$Qu$.push($Gn$)
**while** $Qu$ is not empty **do**
  $L \leftarrow Qu$.dequeue()
  **if** $L$ does not exist in $Kit$ **then**
    $C \leftarrow$ Find all functional units that create $L$
    Let max = -1
    **for** each cand in C **do**
      If cand.successRate > max then
        max = cand.successRate
        $C'$ = cand
        $C' \leftarrow$ Select one from $C$ with a max value
    **end for**
    $T$.append($C'$)
    **for** each input n in $C'$ **do**
      **if** n is not visited **then**
        $Qu$.enqueue(n)
        Mark n as visited
      **end if**
    **end for**
  **end if**
**end while**
$T$.reverse()
**Output**: $T$

---

*Heuristic 2:* This algorithm is similar to the second algorithm but while selecting the candidate unit the number of input nodes and their ingredients is considered. A functional unit with a minimum no of input nodes will be selected as a candidate unit at every level.

---
Algorithm 3: Task tree retrieval using heuristic 2
---

**Input**: reference goal node $Gn$ and given ingredients $In$
$T \leftarrow$ A list of functional units in the task tree
$Kit \leftarrow$ A list of items available in the kitchen
$Qu \leftarrow$ A queue of items to search
$Qu$.push($Gn$)
**while** $Qu$ is not empty **do**
  $L \leftarrow Qu$.dequeue()
  **if** $L$ does not exist in $Kit$ **then**
    $C \leftarrow$ Find all functional units that create $L$
    h2 = {} // dictionary to maintain fu inputs count
    **for** each cand in C **do**
      h2[cand] = cand.inputs_count
    **end for**
    $C'$ = h2.k for k, y in h2 then min(h2.y)
    $C' \leftarrow$ Select one from $C$ with a minimum(h2)
    $T$.append($C'$)
    **for** each input n in $C'$ **do**
      **if** n is not visited **then**

```
        Qu.enqueue(n)
        Mark n as visited
      end if
    end for
  end if
end while
T.reverse()
Output: T
```
---

## IV. DISCUSSION

Iterative deepening search traverses the FOON by performing DFS and Bfd at the given depth bound. It will keep on increasing the depth level until the solution is found. If the solution presents at a deeper level this algorithm takes more time to fetch the task tree. For every new depth-bound increment this will traverse all the previously traversed nodes which increase the time complexity. As Heuristics 1 and 2 follow BFS they easily find the solution at upper levels, but if the solution presents at the deeper levels, then search complexity increases.

| Goal nodes | Iterative deepening search | Heuristic 1 | Heuristic 2 |
|---|---|---|---|
| Greek salad | 28 | 35 | 31 |
| ice | 1 | 1 | 1 |
| macaroni | 7 | 7 | 8 |
| Sweet potato | 3 | 3 | 3 |
| Whipped cream | 10 | 10 | 15 |

Fig. 5: Number of functional units in a task tree for sample goal nodes

All three algorithms may have a task tree containing the same or a different number of functional units. For the goal nodes, ice and sweet potato all task trees have the same number of functional units.

| Goal nodes | Iterative deepening search | Heuristic 1 | Heuristic 2 |
|---|---|---|---|
| Greek salad | 0:00:00.008075 | 0:00:00.007864 | 0:00:00.007998 |
| ice | 0:00:00.001001 | 0:00:00.001002 | 0:00:00.000999 |
| macaroni | 0:00:00.001912 | 0:00:00.002999 | 0:00:00.002009 |
| Sweet potato | 0:00:00 | 0:00:00.001050 | 0:00:00 |
| Whipped cream | 0:00:00.002003 | 0:00:00.002999 | 0:00:00.004043 |

Fig. 6: Time complexity comparison of 3 algorithms for given sample goal nodes